\pdfoutput=1
\documentclass[11pt]{article}

\usepackage[preprint]{acl}

\usepackage{times}
\usepackage{latexsym}
\usepackage{xcolor}
\usepackage[normalem]{ulem}

\usepackage{times}
\usepackage{latexsym}
\usepackage{hyperref}
\usepackage[T1]{fontenc}
\usepackage[utf8]{inputenc}
\usepackage{microtype}
\usepackage{inconsolata}
\usepackage{graphicx}
\usepackage{listings}
\usepackage[T1]{fontenc}
\usepackage[utf8]{inputenc}

\usepackage{microtype}
\usepackage{booktabs}
\usepackage{balance}

\usepackage{inconsolata}

\usepackage{graphicx}

\title{Adversarial Creation and Detection of AI-Generated Social Bot Content}

\author{
 \textbf{Mykola Trokhymovych\textsuperscript{1,2}},
 \textbf{Ricardo Baeza-Yates\textsuperscript{1,3}},
 \textbf{Alessandro Flammini\textsuperscript{2}},
 \\
 \textbf{Diego Saez-Trumper\textsuperscript{1}},
 \textbf{Filippo Menczer\textsuperscript{2}}
\\
\\
 \textsuperscript{1}Universitat Pompeu Fabra, Barcelona, Spain,
 \\
 \textsuperscript{2}Observatory on Social Media, Indiana University, Bloomington, Indiana, USA,
 \\
 \textsuperscript{3}KTH Royal Institute of Technology, Stockholm, Sweden
\\
 \small{
   \textbf{Correspondence:} \href{mailto:mykola.trokhymovych@upf.edu}{mykola.trokhymovych@upf.edu}
 }
}

\hypersetup{
  pdftitle={Adversarial Creation and Detection of AI-Generated Social Bot Content},
  pdfauthor={Mykola Trokhymovych, Ricardo Baeza-Yates, Alessandro Flammini, Diego Saez-Trumper, Filippo Menczer},
  pdfkeywords={social bots, AI-generated content, bot detection, large language models, adversarial data generation}
}

\begin{document}
\maketitle

\begin{abstract}
The convergence of large language models and social bots allows malicious actors to manipulate the information ecosystem by generating human-like content at scale. 
Existing models for detecting AI-generated content often fail in the wild, primarily due to the lack of ground-truth data. 
We address this gap through an adversarial methodology that models the impersonation of real social media users by malicious actors. 
Using this methodology, we curate a multilingual, cross-platform dataset of paired human and AI-generated messages. 
Training on such adversarial data yields accurate detection of AI-generated text. 
Our approach significantly outperforms existing models for content-based bot detection in real-world, out-of-distribution data.  
\end{abstract}

\section{Introduction}

The convergence of Large Language Models (LLMs) and social bots enables the generation of inauthentic content and interactions at scale, for example to spread misinformation on social media~\cite{mozes2023usellmsillicitpurposes}.
This poses unprecedented threats to democracy~\cite{Kunst25AIswarms}. 
While bot detection tools have historically relied on metadata and network analysis in combination with basic content analysis~\cite{socialbots-CACM,Yang2022Botometer101,SocialBots24BookChapter}, those methods are ineffective at detecting sophisticated bots that employ AI models~\cite{Yang_2024}. We posit that advanced content analysis may provide stronger clues about AI-supported bots.

However, it is now challenging to distinguish AI-generated content from human text~\cite{FIEDLER2025100321}. OpenAI, for instance, withdrew its classification tool because it could not reliably spot the difference.\footnote{\url{https://openai.com/blog/new-ai-classifier-for-indicating-ai-written-text}}
Recent advancements propose various solutions, ranging from supervised~\cite{wang-etal-2024-m4} and zero-shot~\cite{10.5555/3692070.3692768} models to retrieval~\cite{sadasivan2024aigeneratedtextreliablydetected} and watermarking~\cite{kirchenbauer2024watermarklargelanguagemodels} techniques.

\begin{figure}
  \centering
  \includegraphics[width=1\linewidth]{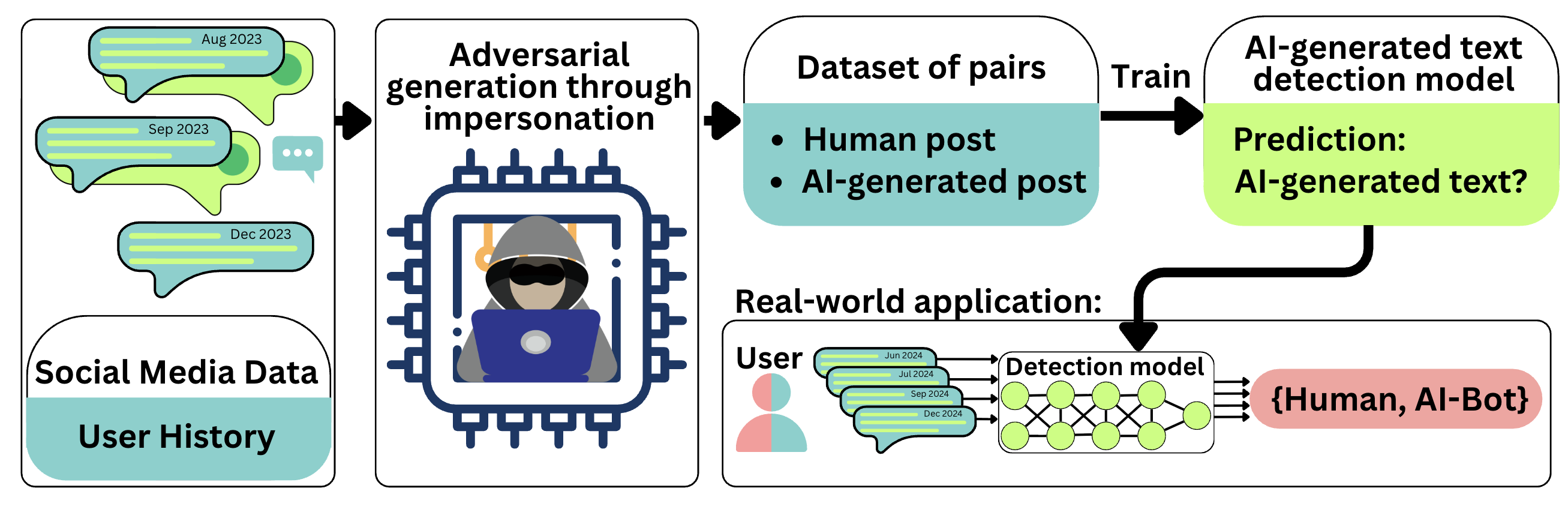}
  \caption{Pipeline for content-based detection of AI-powered social bots.}
  \label{fig:teaser}
\end{figure}

Almost all of these methods suffer from a critical limitation: they perform poorly with short text~\cite{chakraborty2023possibilitiesaigeneratedtextdetection}. 
This is particularly concerning given that social media posts are usually short and stylistically diverse.
Furthermore, there is a lack of robust benchmarks that capture the complexity of AI-generated text in social media. 
Most existing datasets rely on generic paraphrasing or direct generation strategies rather than realistic user imitation~\cite{macko-etal-2025-multisocial}. 

A way to generate realistic training data for the robust detection of AI-powered bots is to emulate such bots in an adversarial setting. 
To this end, we constructed a data-generation pipeline that attempts to capture the behavior of potential malicious actors. 
Specifically, our methodology is to imitate real users who write to specific discussions, based on their profiles and historical messaging behaviors. 
Rather than treating the detection of AI-generated social media content purely as a text classification problem, this approach captures crucial contextual dimensions like platform affordances and the identity, interactions, stance, and unique writing style of the emulated content creator.  

Our pipeline shown in Figure ~\ref{fig:teaser} constructs a dataset of paired human and AI-generated messages, enriched with social network metadata, which presents a challenge for existing detectors. 
We leverage this dataset to train new classification models and demonstrate their effectiveness in detecting AI-powered social bots on real-world, out-of-distribution data.

The key contributions of this study are two-fold: 
\begin{itemize}  
\item An adversarial methodology that mimics real users by generating social media content; and
\item Content-based models to detect AI-powered social bots with high accuracy on out-of-distribution, real-world data. 
\end{itemize}

The rest of the paper is organized as follows. Section 2 reviews related work on automated social media behavior and AI-generated content detection. Section 3 describes the data collection, generation pipeline, and dataset characteristics. Section 4 presents the detection models, including both training-free and training-based approaches. Section 5 reports experimental results and sensitivity analysis. Section 6 concludes the paper and discusses limitations and ethical considerations.

\section{Related Work}

\subsection{Automated Behavior in Social Media}

Computer algorithms that automatically produce content and interact with humans on social media (a.k.a.~social bots) have long been identified as influential actors online~\cite{socialbots-CACM,Shao18hoaxybots,Benevenuto2010DetectingSO}.

Historically, social bots have been characterized by behavioral features such as high posting rates, regular activity, and anomalous network connectivity strategies~\cite{6280553,Varol_Ferrara_Davis_Menczer_Flammini_2017}. Such bots often use easily-automated linguistic patterns (e.g., heavy use of hashtags, repetitive positive terms) and simple replies, in contrast to the more conversational style of human users~\cite{Ng2025AGC}.

Detection frameworks like Botometer have operationalized these signals, analyzing features extracted from account metadata, behavioral patterns, social network structure, and content to estimate the likelihood that an account is a bot~\cite{10.1145/2872518.2889302,Yang2022Botometer101}. At the same time, bot operators actively improve their strategies to bypass detection models, creating an arms race that requires frequent tool refinement~\cite{https://doi.org/10.1002/hbe2.115}.

This landscape has shifted dramatically in recent years. 
On the one hand, platforms are making it difficult to access data for extracting account features beyond content. 
On the other hand, the rise of generative AI for content generation now allows bots to produce human-like text that makes traditional content-based detection algorithms ineffective~\cite{Ferrara_2024,Yang_2024}.

\subsection{AI-Generated Content in Social Media}
\label{sec:2.2}

The rapid development of large language models has substantially increased the realism of synthetic text~\cite{10.1145/3624725}. Modern LLMs can produce coherent, human-like content adapted for specific scenarios using only simple prompts, without the need for additional training~\cite{10.5555/3495724.3495883}. The availability of open-weight models and inference-as-a-service platforms has substantially boosted the accessibility of these tools in recent years~\cite{wolf-etal-2020-transformers}. 

These models can be used to simulate social media personas and participate in online conversations by imitating the stylistic nuances of authentic user text~\cite{malik-etal-2024-empirical,balog2024realisticsyntheticusergeneratedcontent}. A recent study found a substantial increase in the rate of AI-generated text on social media since 2022, when LLMs became widely used by the general public~\cite{sun-etal-2025-ai}. 

Detecting AI-generated content is important to understand its impact in general, and in particular to combat malicious applications such as impersonation, fraud, fake reviews, and disinformation~\cite{AI-faces,10177704,10.1145/3531146.3533088}. 
However, the lack of training data that accurately reflect real-world AI usage is a critical challenge to building robust detection systems. 
Although numerous prior studies have contributed datasets of AI-generated content, few have focused on social media. Selected examples of existing resources include MGTBench (essays, news, Reddit stories)~\cite{HSCBZ24}, M4GT-Bench (multi-domain, with limited social media coverage and restricted to English)~\cite{wang-etal-2024-m4gt}, MULTITuDE (news only)~\cite{Macko_2023}, and MAiDE-up (hotel reviews)~\cite{ignat-etal-2025-maide}.  

AIGTBench~\cite{sun-etal-2025-ai} aggregates human and synthetic text. The latter was generated by polishing, question-answering, and summary expansion based on articles from publishing and social media platforms. However, AIGTBench includes only English and has limited personalization. In contrast, the MultiSocial dataset~\cite{macko-etal-2025-multisocial} offers a multilingual and multi-platform corpus, where synthetic samples were primarily generated by rephrasing social media messages. 

These resources rely exclusively on synthetic AI-generated data, created in a controlled setting. 
Conversely, the Fox8-23 dataset~\cite{Yang_2024} employs a collection strategy in the wild. The authors curated content from active Twitter accounts, establishing ground truth by identifying AI-powered social bots through their self-revealing messages.  

In this paper, we also utilize a controlled generation setting. Unlike prior work, we incorporate social media context and imitate real user writing styles by conditioning the generation on a user's persona and past messages.

\section{Data}
\label{sec:data_collection}

\begin{figure*}
  \centering
  \includegraphics[width=0.9\linewidth]{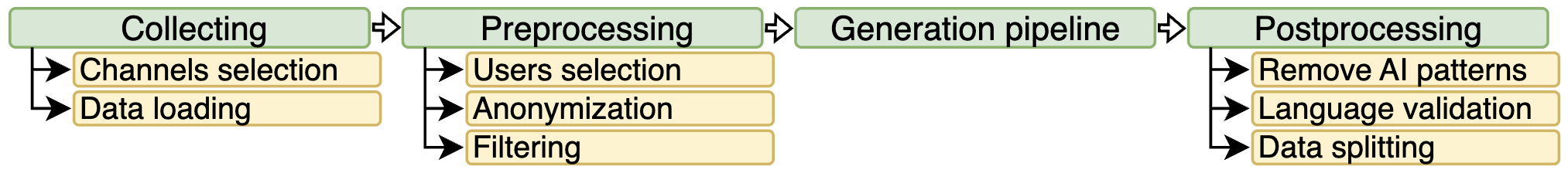}
  \caption{Diagram of dataset curation steps.}
  \label{fig:data_collection_scheme}
\end{figure*}

\begin{figure*}
  \centering
  \includegraphics[width=0.9\linewidth]{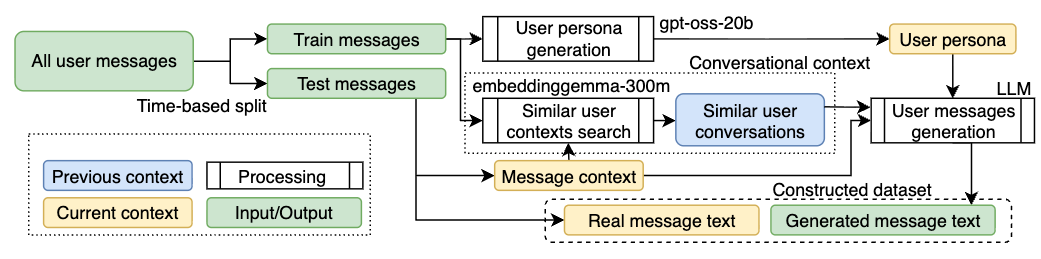}
  \caption{Pipeline for creating realistic AI-generated messages in a social media context.}
  \label{fig:generation_pipeline}
\end{figure*}

In this section, we present the methodology for constructing a dataset of paired human and AI-generated messages (see Figure~\ref{fig:data_collection_scheme}). We start from real-world social media conversations, subsequently enriched with artificially generated messages. We describe the rationale for curating the initial data, the pipeline used to extend it with AI-generated content, and the key properties of the resulting dataset.

\subsection{Data Collection}

We collect a multilingual dataset that includes a variety of writing systems (e.g., Latin, Cyrillic, Arabic). 
We also ensure the inclusion of both high-resource and low-resource languages, at least for the testing part of the dataset. 
We use two large-scale communication platforms, Telegram and Reddit. 

We collect Reddit data using the ConvoKit tool, which allows access to posts and comments created until October 2018~\cite{baumgartner2020pushshiftredditdataset,chang-etal-2020-convokit}. Reddit is organized into subreddits, each typically focused on a specific topic. Although most of the subreddits are in English, there are communities where users communicate in other languages. These non-English subreddits are often country-specific and cover a broad range of topics, e.g., r/bulgaria or r/ukraina~\cite{10.1145/3447535.3462481}. We use this structure as a proxy for identifying conversations in non-English languages, assuming that subreddits dedicated to a particular country mostly contain content in the national language. For English data, we select subreddits focused on finance and politics to ensure topical diversity. Additionally, for some languages, we include subreddits from multiple countries to capture potential regional variation (e.g., r/es and r/chile for Spanish; r/portugal and r/brasil for Portuguese). We collect data from 20 subreddits that cover 15 languages.

For Telegram, we export the full chat history using the official Telegram desktop application. To select channels, we use the Telemetrio website,\footnote{\url{https://telemetr.io/}, accessed 01-04-2025} focusing on public channels that provide access to open chat histories and that have the highest subscriber counts in the news or politics categories. At the time of data collection (10 April 2025), we downloaded the full chat history available for each selected channel. We gathered data from 16 open chats that cover 13 languages. 
In total, the dataset covers 17 languages.  

\subsection{Data Processing}

We convert the data from both Telegram and Reddit conversations into a common thread format. Each thread consists of an initial post followed by its subsequent discussion. Only textual content is processed; all media files are excluded to avoid storing copyrighted or potentially illegal content, in line with prior practice~\cite{10.1145/3690624.3709397}. Users are identified by their nicknames, which we anonymize using randomized identifiers to protect user privacy. We also replace any known nicknames appearing in messages with their corresponding anonymous identifiers.

For each channel or subreddit, we randomly sample up to 200 users, restricting our selection to those who have participated in at least 15 threads. This criterion ensures sufficient conversational history per user, which is necessary for the AI-based message generation process described later. In cases where fewer than 200 eligible users are available, all qualifying users are included. This sampling strategy helps balance the dataset across different languages and topics, contributing to more representative and diverse data coverage. 
Messages from Reddit users whose original nicknames appear as \textit{[removed]} or \textit{[deleted]} are excluded from the dataset, as we treat these cases as deleted content. 

\subsection{Generation Pipeline}

Message generation follows a structured, multi-step pipeline (see Figure~\ref{fig:generation_pipeline}). 
The goal of this process is to generate realistic messages by imitating the behavior of a specific user within a given thread. 
The process includes three main components: (i)~constructing a user persona based on their historical messages; (ii)~retrieving conversational context, i.e., threads that are semantically similar to the current one to provide information about how the user typically responds in comparable situations; and (iii)~generating the final message by prompting a language model to respond within the given thread.

First, the threads associated with each user are split into training and testing sets based on timestamps. The testing set, used for generating AI message pairs, consists of the most recent half of the user's threads, capped at a maximum of 20 threads per user. The remaining threads are used for constructing the user persona and conversational context. 

The user persona is generated by prompting a language model to produce a brief user description, identify the languages used in communication, and determine topics associated with both positive and negative sentiment. To construct the prompt, we sample up to 10 random conversations from the user's training data and provide them as input. For this task, we use the \textit{openai/gpt-oss-20b} model, configured to produce a structured output format suitable for downstream use. Full details of the prompt and model parameters are provided in Appendix~\ref{app:prompting}. 

We generate text messages in two modes: with and without conversational context. 
Incorporating this context into the prompt allows us to more directly reflect the user's writing style, going beyond a generic user persona, which often struggles to fully capture the complexities of individual experiences and communication nuances~\cite{malik-etal-2024-empirical,ng2024llmsechousevaluating}. 
To retrieve related threads for the conversational context, we use the \textit{google/embeddinggemma-300m}\footnote{\url{https://huggingface.co/google/embeddinggemma-300m}} model to generate embeddings for each thread and perform similarity search using cosine similarity. This model was selected for its strong performance on benchmark tasks, robust multilingual support, and extended context window capabilities. For each target thread in the test set, we retrieve the five most similar threads from the user's training data, which are then used as additional context during the final message generation stage. 

Finally, the target message context, the generated user persona, and optionally the conversational context are provided as input to a generative model instructed to imitate the target user. 
Specifically, the model generates a response to the thread in the same position where the target user did so. 
For this step, we use two open-source instruction-tuned models from different providers and with different numbers of parameters: \textit{Gemma-3n-E4B (8B)}\footnote{\url{https://huggingface.co/google/gemma-3n-E4B-it}} and \textit{Qwen3-235B-A22B (235B)}.\footnote{\url{https://huggingface.co/Qwen/Qwen3-235B-A22B-Instruct-2507}} 
Full details of the prompt structure and model parameters are provided in Appendix~\ref{app:prompting}. 

To increase the difficulty of distinguishing between real and AI-generated messages, we explicitly instruct each model to generate responses with a length approximately matching the original human-written message; messages with fewer than ten characters of text are excluded. Furthermore, we consider only the first message authored by the target user within each thread as the generation target, discarding any subsequent messages in the same thread.

\subsection{Data Postprocessing}
\label{sec:data_processing}

To make the dataset more realistic and challenging, we replace long dashes with short dashes and curly quotation marks with straight quotes in the generated texts, unless the target user used such characters in their prior conversations. 
This adjustment is intended to eliminate basic textual artifacts that may signal AI-generated content~\cite{das2024surfacetrackingartifactualityllmgenerated}. 
We avoid advanced humanization techniques or stylistic enhancements to keep the generation scenario simple and still realistic. 
We also filter out approximately 0.2\% of samples that include any artifacts related to the prompt.

We estimate the language of both real and generated messages using the \textit{lingua-py} tool.\footnote{\url{https://github.com/pemistahl/lingua-py}} 
We filter out pairs in which the generated text does not match the language of the real one. 
We further filter out pairs where the detected language of the real or generated text is undefined. This often happens when text is too short or consists mostly of emoticons, numbers, symbols, or gibberish. 
We also filter pairs where the language is not among the 17 expected languages, as these are outside our scope.

Finally, we split the data at the user level to avoid data leakage between the training and testing sets. For each communication channel, we randomly select either 25\% of users or a minimum of 50 users for smaller channels. Additionally, we reserve a 5\% random sample of the training data as a validation set, applying the same user-based splitting logic to ensure consistency.

\subsection{Data Characteristics}

The dataset includes 73,521 unique real user messages, created in 36 Reddit or Telegram channels by 6,326 unique users in 17 languages. Considering the two generative models and two conversational context conditions, the dataset comprises 263,594 pairs of real and generated text. The testing part of the dataset comprises 1,772 unique users, with 71,455 messages generated in 14,288 unique threads.

Initial data analysis shows statistical differences between real and generated messages. On average, generated text is shorter (136 vs.\ 156 characters). 
Messages generated with conversational context are shorter than those without (130 vs.\ 143 characters). 
Moreover, real text contains a significantly higher density of links (4.2\% vs.\ 0.6\%) and user mentions (0.46\% vs.\ 0.29\%).
These differences decrease when context is provided; for instance, the link rate for text generated without conversational context is 0.3\% compared to 0.9\% for those with context. 
This suggests that context integration results in generated content that more closely imitates the style of real users.  

\section{Detection Models}
\label{sec:4}

With the dataset of social-media text generated by imitating real users, we proceed to analyze how difficult it is to distinguish AI-generated messages from their corresponding original ones. Following the taxonomy of LLM-generated content detection models by \citet{yang-etal-2024-survey}, we test training-free (a.k.a. zero-shot) detection and training-based methods. 
Each method produces a numerical score used to classify a message.

\subsection{Training-Free Detectors}

Training-free detectors rely on statistical patterns in text to distinguish between human and AI writing. This makes them broadly applicable and less dependent on the specific generation model, compared to training-based methods~\cite{10.1162/coli_a_00549}. 
This robustness is especially valuable for social-media data, where numerous users contribute diverse writing styles~\cite{10.5555/3692070.3692768}.

In our experiments we test different training-free detectors, such as Binoculars~\cite{10.5555/3692070.3692768}, FastDetectGPT~\cite{bao2023fast}, and GECScore~\cite{wu2024GECScore}. 
The Binoculars approach is based on a pair of LLMs. The models used in the original formulation do not support the multiple languages in our dataset, therefore we replace them by a pair of multilingual models of comparable size (Qwen2.5-7B-Instruct and Qwen2.5-7B), following previous research~\cite{quaremba-etal-2025-wetbench}. 
We use the same models for FastDetectGPT. 
As for GECScore, we follow the logic from the original paper but use \textit{gpt-5-nano-2025-08-07}, a more up-to-date and cost-efficient model. 
Additionally, we update the prompts to follow industry best practices, providing clearer instructions (see Appendix~\ref{app:prompting}).

\subsection{Training-Based Detectors}

Training-based models generally achieve significantly better accuracy in detecting AI-generated social media content, but also tend to overfit, limiting generalization~\cite{macko-etal-2025-multisocial}. 
Using our dataset, we train custom detection models and evaluate both accuracy and generalization.
We test two modeling approaches: 
\begin{itemize}
    \item \textbf{LFC}: a linguistic features classifier;
    \item \textbf{TC}: a transformer-based classifier.
\end{itemize}

For the LFC, we first use the LFTK tool\footnote{\url{https://github.com/brucewlee/lftk}} to extract the set of handcrafted linguistic features from text~\cite{lee-lee-2023-lftk}. We then train a classification model based on gradient boosting~\cite{DBLP:journals/corr/DorogushGGKPV17}.

For the TC, we experiment with three multilingual base models: mBERT,\footnote{\url{google-bert/bert-base-multilingual-cased}} XLM-RoBERTa,\footnote{\url{https://huggingface.co/FacebookAI/xlm-roberta-base}} and Gemma 3 1B.\footnote{\url{https://huggingface.co/google/gemma-3-1b-pt}}  We fine-tune each model for the classification task. Training details are explained in Appendix~\ref{app:training}.

\subsection{Evaluation}

Model accuracy is primarily assessed using the Area Under the Curve (ROC-AUC) metric, which quantifies the model's ability to discriminate between human-authored and machine-generated messages across all classification thresholds. 
We use bootstrapping to compute confidence intervals (see Appendix~\ref{sec:confidence_intervals} for details). 
To evaluate the robustness and generalization of the proposed models, we utilize Fox8-23 (cf.~\S~\ref{sec:2.2}) as an independent benchmark. 

Finally, we average the scores of each user's messages to obtain a user-level score.
The goal is to classify users rather than individual messages, simulating the real-world scenario of detecting AI-powered social bots. 

\section{Results}

\subsection{AI-generated Text Detection}

Firstly, we evaluate the accuracy of models for the task of detecting AI-generated text, where the input is a single social media post. We use the testing hold-out dataset and evaluate 
the training-free and training-based detectors discussed in \S~\ref{sec:4}. 
In addition, we evaluate two other supervised models trained on other datasets: 
OSM-Det\footnote{Training data was restricted to English; results may not generalize to multilingual contexts.} \cite{sun-etal-2025-ai} and a transformer-based classifier based on XLM-RoBERTa, trained on the MultiSocial dataset (MS-based TC).

\begin{table}
\begin{center} 
  {\tabcolsep=1.4mm
  \begin{tabular}{c|cc}
    \hline
    \textbf{Model} & \textbf{Our Data} &  \textbf{Fox8-23}\\
        \midrule
        GEC score & 0.646±0.007 & 0.593±0.010 \\
        Binocular & 0.665±0.007 & 0.550±0.011 \\
        FastDetect & 0.681±0.007 & 0.543±0.011 \\
        \midrule
        OSM-Det & 0.603±0.006 & 0.620±0.011 \\
        MS-based TC & 0.700±0.007 & 0.582±0.011 \\
        \midrule
        LFC & 0.858±0.005 & 0.602±0.012 \\
        TC (mBERT) & 0.961±0.002 & 0.720±0.010 \\
        TC (Gemma3) & 0.964±0.002 & \textbf{0.722±0.010} \\
        TC (RoBERTa) & \textbf{0.969±0.002} & 0.714±0.010 \\
        \bottomrule
  \end{tabular}
  \caption{AUC for the task of detecting AI-generated text. 95\% confidence intervals are shown in this and the following tables. The best model is shown in bold.}
    \label{tab:our_dataset_full}
  }
\end{center}
\end{table}

\begin{table}
\small
\begin{center} 
  {\tabcolsep=0.8mm
  \begin{tabular}{c|ccc}
    \hline
    \textbf{Model} & \textbf{MultiSocial} &  \textbf{AIGT} & \textbf{Fox8-23}\\
        \midrule
        Binocular & 0.709±0.015 & 0.740±0.010 & 0.550±0.011 \\
        \midrule
        OSM-Det & 0.675±0.015 & \textbf{0.999±0.001} & \textbf{0.620±0.011} \\
        MS-based TC & \textbf{0.974±0.004} & 0.935±0.005 & 0.582±0.011 \\
        \bottomrule
  \end{tabular}
    \caption{AUC for the detection of AI-generated text in external datasets.}
    \label{tab:external_dataset_full}
  }
\end{center}
\end{table}

The results in Table~\ref{tab:our_dataset_full} demonstrate that models trained on our data significantly outperform all baselines.  
The TC model based on the XLM-RoBERTa encoder is the most accurate. 
Training-free classifiers achieve accuracy comparable with models trained on external datasets.

To properly assess real-world robustness, we prioritize the Fox8-23 dataset. This is the only dataset strictly unseen during the training of all evaluated models and consists of ``in-the-wild'' data, representing a highly realistic application scenario. 
Table~\ref{tab:external_dataset_full} reveals a consistent pattern of in-domain bias among external models: the MultiSocial-based TC and OSM-Det achieve near-perfect scores on their respective training distributions, yet their accuracy drops substantially on Fox8-23. 
This suggests that generation strategies relying on rephrasing or modifying existing content do not provide sufficient training signal for robust, real-world detection. In contrast, models trained on our data significantly outperform all baselines on this benchmark, demonstrating the value of realistic, context-aware adversarial data.

\subsection{AI-Powered Social Bot Detection}

\begin{table}
\begin{center} 
  {\tabcolsep=0.8mm
  \begin{tabular}{c|cc}
    \hline
    \textbf{Model} & \textbf{Our Data} &  \textbf{Fox8-23}\\
    \midrule
        GEC score & 0.726±0.006 & 0.811±0.009 \\
        Binocular & 0.754±0.005 & 0.688±0.011 \\
        FastDetect & 0.797±0.005 & 0.672±0.012 \\
        \midrule
        OSM-Det & 0.626±0.005 & 0.861±0.008 \\
        MS-based TC & 0.766±0.006 & 0.744±0.011 \\
        \midrule
        LFC & 0.960±0.003 & 0.907±0.007 \\
        TC (mBERT) & 0.991±0.001 & \textbf{0.989±0.002} \\
        TC (Gemma3) & 0.993±0.001 & 0.941±0.006 \\
        TC (RoBERTa) & \textbf{0.993±0.001} & 0.972±0.004 \\
        \bottomrule
  \end{tabular}
  \caption{AUC for the detection of AI-powered social bots.}
    \label{tab:bots_detection}
  }
\end{center}
\end{table}

Our findings show that detecting AI-generated content at the message level is challenging, especially given the moderate accuracy on out-of-distribution data. However, real-world setups usually require detecting the users who post AI-generated content (AI-powered social bots) rather than classifying each message independently. To simulate such a scenario, we evaluate accuracy at the user level. 

We employ the Fox8-23 dataset as the primary out-of-distribution benchmark for this scenario. As mentioned earlier, this dataset consists of ``in-the-wild'' data. 
It includes user identifiers, making it possible to evaluate our models on this user classification task. 

Our results are shown in Table~\ref{tab:bots_detection}. 
All models achieve significantly higher AUC scores for detecting AI-powered social bots than for individual message classification. 
The accuracy gain is observed on both our synthetically generated dataset and the real-world data (Fox8-23), but more pronounced on the latter. 
Transformer-based classifiers trained on our dataset significantly outperform other configurations. The best model achieves near-perfect accuracy on the Fox8-23 benchmark.

\begin{figure}
  \centering
  \includegraphics[width=1\linewidth]{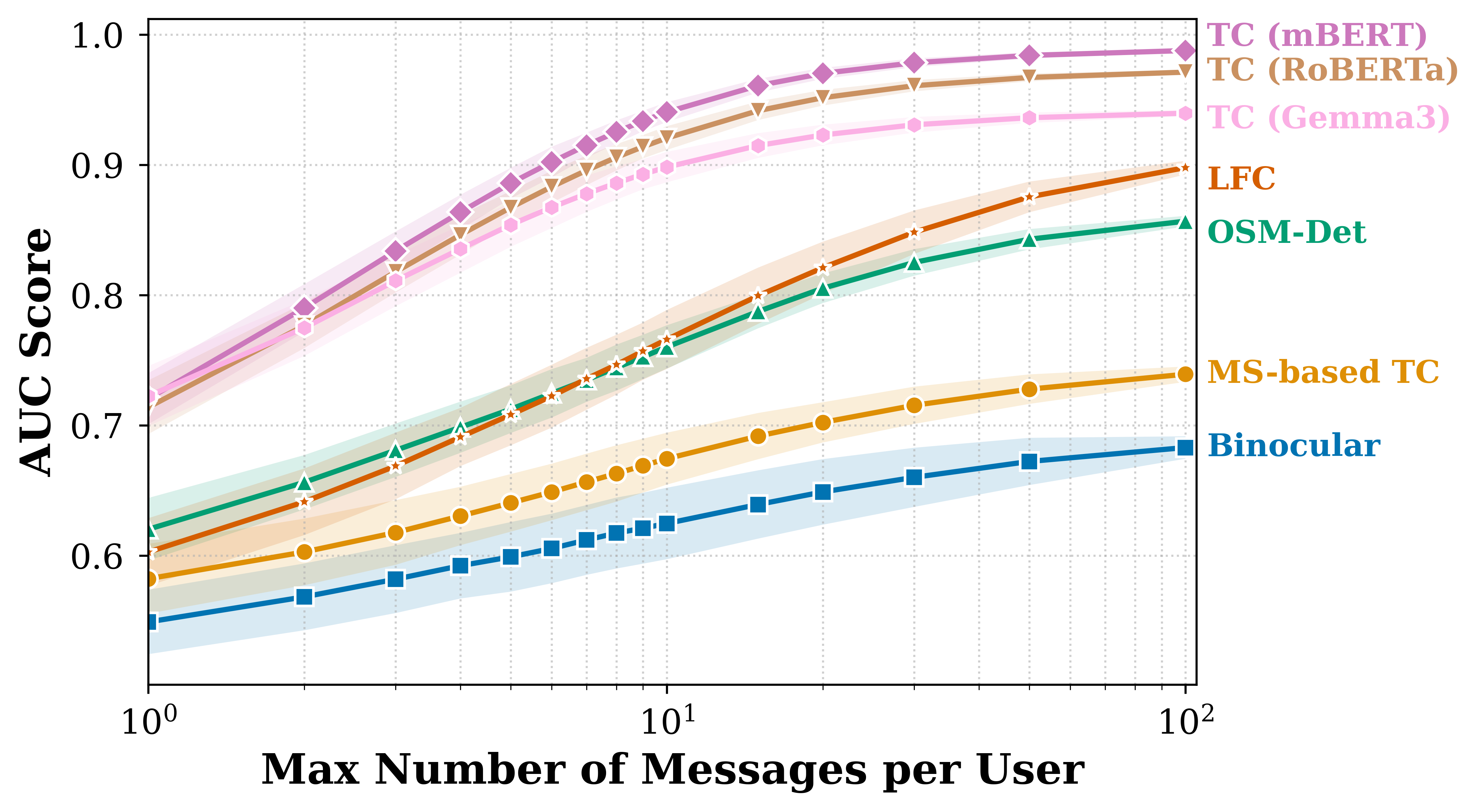}
  \caption{AI-Powered bot detection accuracy as a function of the number of messages $N$ per user, based on the Fox8-23 dataset. For each user, we randomly sampled $N$ messages without replacement, calculating the mean AUC and a 95\% confidence interval across 100 iterations per data point. 
}
  \label{fig:n_messages}
\end{figure}

Figure \ref{fig:n_messages} reports on the impact of the number of messages available per user on detection accuracy, using the Fox8-23 dataset. 
As expected, AUC increases with the number of messages; with just 20 messages per user, our TC (mBERT) model achieves an AUC of approximately 0.97.

\subsection{Sensitivity Analysis}
\label{sec:sensitivity_analysis}

\begin{table}
\small
\begin{center} 
  {\tabcolsep=0.8mm
  \begin{tabular}{c|ccc}
    \hline
    \textbf{Model} & \textbf{0–50} &  \textbf{50–150} & \textbf{150+}\\
        \midrule
        GEC score & 0.593±0.007 & 0.663±0.007 & 0.722±0.007 \\
        Binocular & 0.615±0.007 & 0.671±0.007 & 0.780±0.006 \\
        FastDetect & 0.620±0.007 & 0.673±0.007 & 0.781±0.006 \\
        \midrule
        OSM-Det & 0.577±0.006 & 0.643±0.006 & 0.591±0.007 \\
        MS-based TC & 0.632±0.007 & 0.686±0.007 & 0.786±0.006 \\
        \midrule
        LFC & 0.772±0.006 & 0.854±0.005 & 0.926±0.004 \\
        TC (mBERT) & 0.907±0.004 & 0.967±0.002 & 0.987±0.001 \\
        TC (Gemma3) & \textbf{0.922±0.004} & 0.966±0.002 & 0.987±0.002 \\
        TC (RoBERTa) & 0.920±0.004 & \textbf{0.976±0.002} & \textbf{0.990±0.001} \\
        \bottomrule
  \end{tabular}
      \caption{Model AUC for the task of AI-generated text
detection depending on the length of the text (in number of characters).}
    \label{tab:length}
  }
\end{center}
\end{table}

\begin{table}
\begin{center} 
  {\tabcolsep=0.8mm
  \begin{tabular}{c|cc}
    \hline
    \textbf{Model} & \textbf{Reddit} &  \textbf{Telegram}\\
        \midrule
        GEC score & 0.640±0.007 & 0.654±0.007 \\
        Binocular & 0.655±0.007 & 0.679±0.006 \\
        FastDetect & 0.672±0.007 & 0.694±0.007 \\
        \midrule
        OSM-Det & 0.581±0.007 & 0.636±0.006 \\
        MS-based TC & 0.688±0.007 & 0.723±0.006 \\
        \midrule
        LFC & 0.877±0.005 & 0.835±0.006 \\
        TC (mBERT) & 0.974±0.002 & 0.942±0.003 \\
        TC (Gemma3) & 0.974±0.002 & 0.950±0.003 \\
        TC (RoBERTa) & \textbf{0.980±0.002} & \textbf{0.954±0.003} \\
        \bottomrule
  \end{tabular}
    \caption{Model AUC for the task of AI-generated text
detection depending on the platform.}
    \label{tab:platform}
  }
\end{center}
\end{table}

\begin{table}
\begin{center} 
  {\tabcolsep=0.8mm
  \begin{tabular}{c|cc}
    \hline
    \textbf{Model} & \textbf{Gemma} & \textbf{Qwen}\\
        \midrule
         GEC score & 0.678±0.007 & 0.617±0.007 \\
        Binocular & 0.727±0.007 & 0.607±0.007 \\
        FastDetect & 0.737±0.006 & 0.629±0.007 \\
        \midrule
        OSM-Det & 0.668±0.006 & 0.541±0.006 \\
        MS-based TC & 0.812±0.005 & 0.596±0.007 \\
        \midrule
        LFC & 0.901±0.004 & 0.817±0.006 \\
        TC (mBERT) & 0.970±0.002 & 0.952±0.003 \\
        TC (Gemma3) & 0.970±0.002 & 0.959±0.003 \\
        TC (RoBERTa) & \textbf{0.978±0.002} & \textbf{0.961±0.002} \\
        \bottomrule
  \end{tabular}
     \caption{Model AUC for the task of AI-generated text
detection depending on the model used for generation. Columns are sorted by the size of the model in ascending order.}
    \label{tab:model}
  }
\end{center}
\end{table}

\begin{table}
\begin{center} 
  {\tabcolsep=0.8mm
  \begin{tabular}{c|cc}
    \hline
    \textbf{Model} & \textbf{with context} & \textbf{without context}\\
        \midrule
        GEC score & 0.622±0.007 & 0.671±0.007 \\
        Binocular & 0.650±0.007 & 0.681±0.007 \\
        FastDetect & 0.669±0.007 & 0.693±0.007 \\
        \midrule
        OSM-Det & 0.583±0.006 & 0.624±0.007 \\
        MS-based TC & 0.656±0.007 & 0.747±0.007 \\
        \midrule
        LFC & 0.817±0.006 & 0.901±0.004 \\
        TC (mBERT) & 0.941±0.003 & 0.981±0.001 \\
        TC (Gemma3) & 0.950±0.003 & 0.978±0.002 \\
        TC (RoBERTa) & \textbf{0.953±0.003} & \textbf{0.986±0.001} \\
        \bottomrule
  \end{tabular}
    \caption{Model AUC for the task of AI-generated text
detection depending on the availability of conversational context during message generation.}
    \label{tab:context}
  }
\end{center}
\end{table}

\begin{table*}
\small
\begin{center} 
  {\tabcolsep=0.45mm
  \begin{tabular}{c|ccccccccccccccccc}
    \hline
Model & AR & BG & CA & DE & EL & EN & ES & FA & FR & IT & JA & KO & PL & PT & RU & UK & ZH \\
\midrule
GEC score & 0.651 & 0.702 & 0.695 & 0.695 & 0.675 & 0.631 & 0.682 & 0.589 & 0.706 & 0.677 & 0.407 & 0.739 & 0.653 & 0.640 & 0.620 & 0.642 & 0.443 \\
Binocular & 0.721 & 0.702 & 0.656 & 0.683 & 0.691 & 0.608 & 0.714 & 0.622 & 0.714 & 0.673 & 0.553 & 0.698 & 0.667 & 0.658 & 0.736 & 0.710 & 0.531 \\
FastDetect & 0.734 & 0.720 & 0.681 & 0.699 & 0.716 & 0.623 & 0.728 & 0.634 & 0.724 & 0.695 & 0.600 & 0.693 & 0.688 & 0.676 & 0.749 & 0.726 & 0.565 \\
\midrule
OSM-Det & 0.688 & 0.581 & 0.506 & 0.595 & 0.639 & 0.705 & 0.662 & 0.657 & 0.579 & 0.621 & 0.286 & 0.493 & 0.588 & 0.604 & 0.576 & 0.633 & 0.540 \\
MS-based TC & 0.780 & 0.712 & 0.733 & 0.767 & 0.756 & 0.655 & 0.712 & 0.760 & 0.745 & 0.708 & 0.589 & 0.760 & 0.712 & 0.691 & 0.683 & 0.752 & 0.440 \\
\midrule
LFC & 0.832 & 0.818 & 0.859 & 0.866 & 0.810 & 0.909 & 0.837 & 0.825 & 0.880 & 0.876 & 0.837 & 0.760 & 0.836 & 0.859 & 0.831 & 0.844 & 0.716 \\
TC (mBERT) & 0.928 & 0.965 & 0.964 & 0.975 & 0.938 & 0.977 & 0.946 & 0.906 & 0.977 & 0.971 & 0.940 & 0.907 & 0.953 & 0.966 & 0.970 & 0.960 & 0.703 \\
TC (Gemma3) & 0.951 & 0.974 & 0.968 & 0.973 & 0.959 & 0.969 & 0.956 & 0.921 & 0.974 & 0.973 & 0.950 & 0.973 & 0.953 & 0.968 & 0.973 & 0.971 & 0.725 \\
TC (RoBERTa) & 0.946 & 0.978 & 0.978 & 0.980 & 0.958 & 0.978 & 0.957 & 0.932 & 0.984 & 0.979 & 0.941 & 0.951 & 0.961 & 0.973 & 0.976 & 0.970 & 0.733 \\
  \end{tabular}
  \caption{Model AUC for the task of AI-generated text
detection depending on language.}
    \label{tab:lang}
  }
\end{center}
\end{table*}

We evaluate the sensitivity of our results to various conditions, observing how accuracy is affected by text length, social media platform, the specific model used for generation, availability of conversational context, and language.

Our analysis indicates that detection accuracy is positively correlated with input length across all evaluated architectures (see Table~\ref{tab:length}). Specifically, the top-performing model, TC (RoBERTa), improved its AUC by about 7.5\% as text went from 50 characters to more than 150 characters. 
Similar accuracy gains were observed across other configurations.

No substantial accuracy difference was observed across different social media platforms (see Table~\ref{tab:platform}). The small variations identified may be attributed to differences in average message length; for instance, Reddit messages are typically longer compared to Telegram.

Detection accuracy depends on the size of the LLM used for generation (see Table~\ref{tab:model}). Text produced by smaller models tends to be more detectable than that generated by larger ones. 
Providing conversational context during generation also makes the resulting messages harder to detect for all models (see Table~\ref{tab:context}). 

Finally, language-specific analysis demonstrates that the transformer classifiers trained on our dataset maintain high accuracy across all evaluated languages (see Table~\ref{tab:lang}). 
This suggests that the proposed approach has multilingual capabilities. 
We only observe lower accuracy for Chinese. 

\section{Conclusions}

This paper addresses the growing threat of AI-powered social bots that leverage large language models to generate human-like text. 
Since these bots are defined by the content they produce, detecting them requires analyzing that content directly. 
We proceed from the hypothesis that robust detection of AI-generated social media content requires training on realistic adversarial inputs.

We developed an adversarial data generation pipeline that emulates the behavior of malicious actors, while remaining adaptable to emerging LLMs. 
By conditioning text generation on the historical messaging behavior of users, our approach captures their unique writing styles and stances, producing synthetic content that closely mirrors how real users communicate. 

Our data generation pipeline enabled the construction of a robust, multilingual, and cross-platform dataset comprising paired human and AI-generated messages. 
Using this data, we trained content-based classification models for AI-generated text detection and evaluated them against established baselines. 

Our findings demonstrate that training on realistic, context-aware adversarial data not only achieves high accuracy in detecting AI-generated text, but more importantly, results in substantial accuracy improvement in identifying AI-powered social bots in real-world, out-of-distribution data.

\subsection{Limitations}

We developed our data generation pipeline with a focus on simplicity and scalability, mirroring the strategies likely employed by malicious actors who utilize open-weight LLMs or industry APIs without requiring significant computational resources. 
We acknowledge that more sophisticated strategies exist, such as fine-tuning LLMs for impersonation~\cite{shi2025impersonaevaluatingindividuallevel}. 
These methods demand substantially higher resources and are therefore less likely to be deployed at scale in real-world scenarios. 

Another limitation of this study is the potential presence of existing automated activity in our human baseline, particularly within the Telegram dataset. 
The older Reddit data remains unaffected by recent LLMs. 
Additionally, the scope is limited to two platforms and popular channels. 
The focus on mainstream topics like news and politics introduces a selection bias that may affect the model's accuracy on niche content. 
Future work should incorporate a broader array of platforms and less-prominent discussion topics to mitigate these biases. 

By utilizing two open-weight LLMs of different sizes, we aimed to create a robust and generalizable benchmark. 
We found that larger, more powerful generation models produce content that is harder to detect, leading to a drop in classifier accuracy. 
This highlights a critical requirement for future detection systems: training data must be consistently updated with outputs from state-of-the-art models and varied prompting strategies to follow the evolving capabilities of LLMs. 

Our analysis is limited to textual data. Integrating multi-modal content and activity metadata would likely yield significant improvements in bot detection accuracy. 
Nevertheless, the high-quality signals generated by our current models provide a strong baseline, which can be used in more complex ensemble or hybrid architectures to combat evolving bot threats. 

Finally, we evaluate our approach using the Fox8-23 dataset as a real-world benchmark. 
This dataset has two important limitations. 
First, it was collected before AI content generation had reached its current sophistication. 
Second, it was built by identifying bots that openly revealed themselves, meaning the bots it contains were relatively easy to spot. 
Together, these factors suggest that detection models performing well on this benchmark may struggle significantly against more advanced bots found on today's platforms. 

\subsection{Ethics Statement}

Our models are designed to help identify AI-powered social bots. They can be used as standalone tools or integrated into larger detection systems that include account metadata and behavioral patterns. The primary goal is to provide a reliable signal for researchers and platform moderators to detect AI-generated content. 
These models should not be used for automated account bans. Because models can mistakenly identify real people as bots, any punishing action should involve human review. Furthermore, these models must not be used for the adversarial fine-tuning of LLMs to bypass detection.

By evaluating our models on established, publicly available benchmarks, we ensure reproducibility and adhere to the principles of open science. 
Our data collection methods were reviewed and approved by Indiana University Institutional Review Board. 
To train our models, we used a pre-existing public Reddit dataset from ConvoKit, along with manually exported data from 16 public Telegram broadcast channels and their linked discussion groups. 
We avoided automated scraping to respect the platform infrastructure. 
Detection models must operate in the wild, therefore we have not excluded hateful or inappropriate content that may appear in our training data; we do not endorse this speech. 
We are not publishing our data to avoid compromising the privacy and data ownership of real users. 

While releasing a pipeline for generating realistic AI content presents a dual-use risk, we believe it is a necessary and responsible decision for defensive research.\footnote{\url{https://github.com/trokhymovych/ai-bot-detection}} 
The pipeline utilizes well-known LLM practices, and sharing it enables the research community to actively understand and prepare for existing threats. 
To balance the risk of misuse, we are open-sourcing our trained detection model.\footnote{\url{https://huggingface.co/trokhymovych/mbert-ai-bot-detector}} 

\subsection{Future Work}
Several directions could naturally extend this work. Our generation pipeline relies on prompt-based imitation with open-weight LLMs, mirroring low-resource adversarial setups. Fine-tuning generation models on per-user message histories could achieve closer stylistic imitation~\cite{shi2025impersonaevaluatingindividuallevel}. Incorporating texts generated by specialized models into training corpora is a promising approach to improving detectors for more advanced bots. The dataset can also be extended to include additional platforms and a wider range of topics, which would help mitigate selection bias.

Our detectors currently operate only on message text. Real-world bot detection systems typically combine content signals with account metadata, posting cadence, and network structure~\cite{Yang2022Botometer101,SocialBots24BookChapter}. Studying how to optimally fuse content scores with behavioral and network features is a promising direction. Finally, since larger generation models produce harder-to-detect text (see Section~\ref{sec:sensitivity_analysis}), keeping detection systems effective requires continuous retraining, suggesting the need to periodically refresh training data with outputs from state-of-the-art LLMs and prompting strategies.

\section*{Acknowledgments}
The work of Mykola Trokhymovych is supported by the Google PhD Fellowship and 
MCIN/AEI /10.13039/501100011033 under the Maria de Maeztu Units of Excellence Programme (CEX2021-001195-M).

\bibliography{custom}
\appendix

\section{Additional Modeling Details}
\label{app:training}

\subsection{Model Hyperparameters}

For the linguistic features classifier (LFC), we use the LFTK tool to extract \textit{general}, language-agnostic features. In particular, we use language-specific SpaCy\footnote{\url{https://spacy.io/}} models for languages where they are available. Additionally, we add the detected language as a categorical feature. Later, we use these features to train a CatBoost classification model. Parameters used for training are 5,000 iterations, a 0.01 learning rate, and 500 early stopping rounds based on the accuracy metric on the validation subset.

For the Transformer-based classifier (TC), we experiment with two pretrained base models: mBERT (0.18B parameters) and XLM-RoBERTa (0.28B parameters)~\cite{DBLP:journals/corr/abs-1911-02116,devlin-etal-2019-bert}. 
Both were selected for their multilingual pretraining, which we expect to be beneficial given the nature of the task, while remaining computationally lightweight. 
We fine-tune each model for binary classification over three epochs, using a batch size of 64, a learning rate of $2 \times 10^{-5}$, and a weight decay of 0.01. 
The final checkpoint is selected based on the best accuracy achieved on the validation set during training.

For the Gemma-based TC (1B parameters), a decoder-only model fine-tuned with LoRA ($r{=}16$, $\alpha{=}32$, dropout$=0.1$) targeting query and value projections~\cite{hu2022lora}. We train for one epoch with a learning rate of $2{\times}10^{-4}$, effective batch size of 64 ($4{\times}16$ accumulation steps), and weight decay of 0.01. After training, LoRA weights are merged into the base model for inference.

\subsection{Computational Resources}

All experiments were done on a computational instance powered by a single NVIDIA GB10 Grace Blackwell chip. The experimental environment was based on the NVIDIA container image for PyTorch, Release 25.11. 
In total, approximately 100 GPU hours are required to reproduce the results reported in this paper, excluding the data generation stage, which relies on external API calls.

\section{Confidence Intervals}
\label{sec:confidence_intervals} 

We estimate the confidence intervals for the calculated metrics using bootstrapping~\cite{efron1994introduction}. 
Specifically, we resample 1,000 times by drawing samples of size $N$ with replacement from the testing 
set, where $N$ is the total number of items in the set (capped at 10,000). We report two standard 
deviations as the 95\% confidence interval (CI).

\balance
\section{Data Generation Details}
\label{app:prompting}

We use an OpenAI model solely for persona generation. This is not a key component in the pipeline and can be replaced with other methods, such as manually created personas. The full prompt template used for persona generation is shown in Listing~\ref{prm:persona_prompt}. We use structured output functionality to ensure a unified persona definition within the dataset. In particular, we extract a brief user description, the languages used, and a list of topics with positive and negative attitudes. 

For the final content generation, we use Together.ai as the inference provider for open-source LLMs. In particular, we use two models from different providers and of different sizes. We generate messages in two modes: with and without conversational context. The full prompt templates used for message generation are shown in Listings~\ref{prm:message_generation_prompt_with_context} and~\ref{prm:message_generation_prompt_no_context}. 

The prompt used for GECScore calculation is presented in Listing~\ref{prm:gec_score_prompt}. 

The parameters used for each model are listed below (all parameters not mentioned were left at their default values): 
\begin{itemize}
    \item \textit{openai/gpt-oss-20b}: temperature=1 
    \item \textit{Gemma-3n-E4B}: max\_tokens=2000, temperature=0.5, random\_seed=42
    \item \textit{Qwen3-235B-A22B}: max\_tokens=2000, temperature=0.5, random\_seed=42
\end{itemize}

\begin{lstlisting}[caption={Prompt for GECScore calculation.}, label={prm:gec_score_prompt}]
You are a highly skilled grammar correction AI in multiple languages. You are provided with a text separated by <text></text> tags. Correct any grammatical, spelling, or punctuation errors in the text, doing only minimal changes necessary. Return ONLY the corrected text without any additional commentary.
# Text to correct:
<text>{text}</text>
\end{lstlisting}

\begin{lstlisting}[caption={Prompt template used for user persona generation.}, label={prm:persona_prompt}]
You're an expert in analyzing social media conversations and defining the persona of [USER] based on its interactions.
You will be provided with a set of user interactions in form of conversation under the post, where the user has taken part in ([CONVERSATIONS]).

You need to generate a persona JSON for the user based on the provided interactions.
Pay special attention to users stance to different topics.
The persona JSON should strictly follow the following template: {template_string}
Return ONLY a valid JSON object with exactly these keys:
"description", "languages", "positive_sentiment_topics", "negative_sentiment_topics".
Do not include any explanation or markdown - raw JSON only.
Our target [USER] has ID={anonymized user_id}.
[CONVERSATIONS] {random user threads}
\end{lstlisting}

\begin{lstlisting}[caption={Prompt template used to generate a user message in a given context. This prompt utilizes conversational context that refers to the set of similar previous threads in which the user has participated.}, label={prm:message_generation_prompt_with_context}] 
You are an expert in imitating the user (USER_ID={anonymized user_id}) style in social media communication. 
You are provided: 
- [USER_PERSONA] Characteristics of user usual communication style. 
- [PREVIOUS CONVERSATIONS] User previous conversations. 
- [CURRENT CONTEXT] The current conversation thread where you should write a message on behalf of the user (USER_ID={anonymized user_id}) 
Your task is to imitate the user (USER_ID={anonymized user_id}) and generate the most likely text the user would write in the [CURRENT CONTEXT]. Reproduce from the [PREVIOUS CONVERSATIONS] user's (USER_ID={anonymized user_id}) tone characteristics, tendency to jokes, reacting to posts, etc. 
[USER PERSONA]: {generated user persona} [PREVIOUS CONVERSATIONS]: {user previous conversations} [CURRENT CONTEXT]: {current thread} 
[LENGTH CONSTRAINTS] The response should be strictly around {number of words in real message} words. 
[STYLE CONSTRAINTS]: 
- No Unicode punctuation. Do not beautify or autocorrect.
- You should imitate the user's (USER_ID={anonymized user_id}) style as closely as possible. 
- Use the same language as the user (USER_ID={anonymized user_id}) in the [PREVIOUS CONVERSATIONS]. 
- Output must strictly preserve raw formatting style from [PREVIOUS CONVERSATIONS]. 
- Your response should naturally pick up the conversation from the message with ID {id of the previous message in the thread} within the [CURRENT CONTEXT]. 
[OUTPUT FORMAT] You should return only the response text (your message in thread). Nothing else. Ensure the text style is consistent with user persona. "
\end{lstlisting}

\begin{lstlisting}[caption={Prompt template used to generate a user message in a given context. This prompt does not utilize conversational context.}, label={prm:message_generation_prompt_no_context}] 
You are an expert in imitating the user (USER_ID={anonymized user_id}) style in social media communication.

You are provided: 
- [USER_PERSONA] Characteristics of user usual communication style. 
- [CURRENT CONTEXT] The current conversation thread where you should write a message on behalf of the user (USER_ID={anonymized user_id}) 

Your task is to imitate the user (USER_ID={anonymized user_id}) and generate the most likely text the user would write in the [CURRENT CONTEXT].

[USER PERSONA] {generated user persona} 
[CURRENT CONTEXT]: {current thread}  
[LENGTH CONSTRAINTS] The response should be strictly around {number of words in real message} words.
[STYLE CONSTRAINTS] 
- No Unicode punctuation. Do not beautify or autocorrect.
- You should imitate the user's (USER_ID={anonymized user_id}) style as closely as possible.  
- Your response should naturally pick up the conversation from the message with ID {id of the previous message in the thread} within the [CURRENT CONTEXT]. 
[OUTPUT FORMAT] You should return only the response text (your message in thread). Nothing else. Ensure the text style is consistent with user persona. "
\end{lstlisting}

\end{document}